\pdfoutput=1

\documentclass[11pt]{article}

\usepackage{acl}

\usepackage{times}
\usepackage{latexsym}
\usepackage{booktabs}
\usepackage{multirow}
\usepackage{hyperref}
\usepackage[T1]{fontenc}

\usepackage[utf8]{inputenc}

\usepackage{microtype}
\usepackage{amsfonts}
\usepackage{amsmath}
\usepackage{float}
\usepackage{graphicx}
\usepackage{graphics}

\title{Improving Long Tailed Document-Level Relation Extraction via Easy Relation Augmentation and Contrastive Learning}

\author{Yangkai Du\textsuperscript{1}, Tengfei Ma\textsuperscript{2}, Lingfei Wu\textsuperscript{3}, Yiming Wu\textsuperscript{1}, Xuhong Zhang\textsuperscript{1} \\\bf{Bo Long}\textsuperscript{3}, \bf{Shouling Ji}\textsuperscript{1} \\
        \textsuperscript{1}Zhejiang University; \textsuperscript{2}IBM Research; 
        \textsuperscript{3}JD.COM\\
        \texttt{\{{yangkaidu,wuyiming,zhangxuhong,sji}\}@zju.edu.cn} \\
        \texttt{tengfei.ma1@ibm.com, \{lingfei.wu,bo.long\}@jd.com}
        }

\begin{document}
\maketitle
\begin{abstract}
Towards real-world information extraction scenario, research of relation extraction is advancing to document-level relation extraction(DocRE). Existing approaches for DocRE aim to extract relation by encoding various information sources in the long context by novel model architectures. However, the inherent long-tailed distribution problem of DocRE is overlooked by prior work. We argue that mitigating the long-tailed distribution problem is crucial for DocRE in the real-world scenario. Motivated by the long-tailed distribution problem, we propose an Easy Relation Augmentation(ERA) method for improving DocRE by enhancing the performance of tailed relations. In addition, we further propose a novel contrastive learning framework based on our ERA, i.e., ERACL, which can further improve the model performance on tailed relations and achieve competitive overall DocRE performance compared to the state-of-arts. 

\end{abstract}

\section{Introduction}
Relation extraction plays an essential role in information extraction, which aims to predict relations of entities in texts. Early work on relation extraction mainly focuses on sentence-level relation extraction, i.e., predicting relation from a single sentence, and has achieved promising results. Recently, the research of relation extraction has advanced to document-level relation extraction, a scenario more practical than sentence-level relation extraction and more challenging.

The relation pattern between entity pairs across different sentences is often more complex, and the distance of these entity pairs is relatively long. Therefore, DocRE requires models to figure out the relevant context and conduct reasoning across sentences instead of memorizing the simple relation pattern in a single sentence. Moreover, multiple entity pairs co-exist in one document, and each entity may have more than one mention appearing across sentences. 
Thus, DocRE also requires the model to extract relations of multiple entity pairs from a single document at once. In other words, DocRE is a one-example-multi-instances task while sentence-level RE is a one-example-one-instance task.

\begin{figure}[h]
    \centering
    \includegraphics[width=8cm]{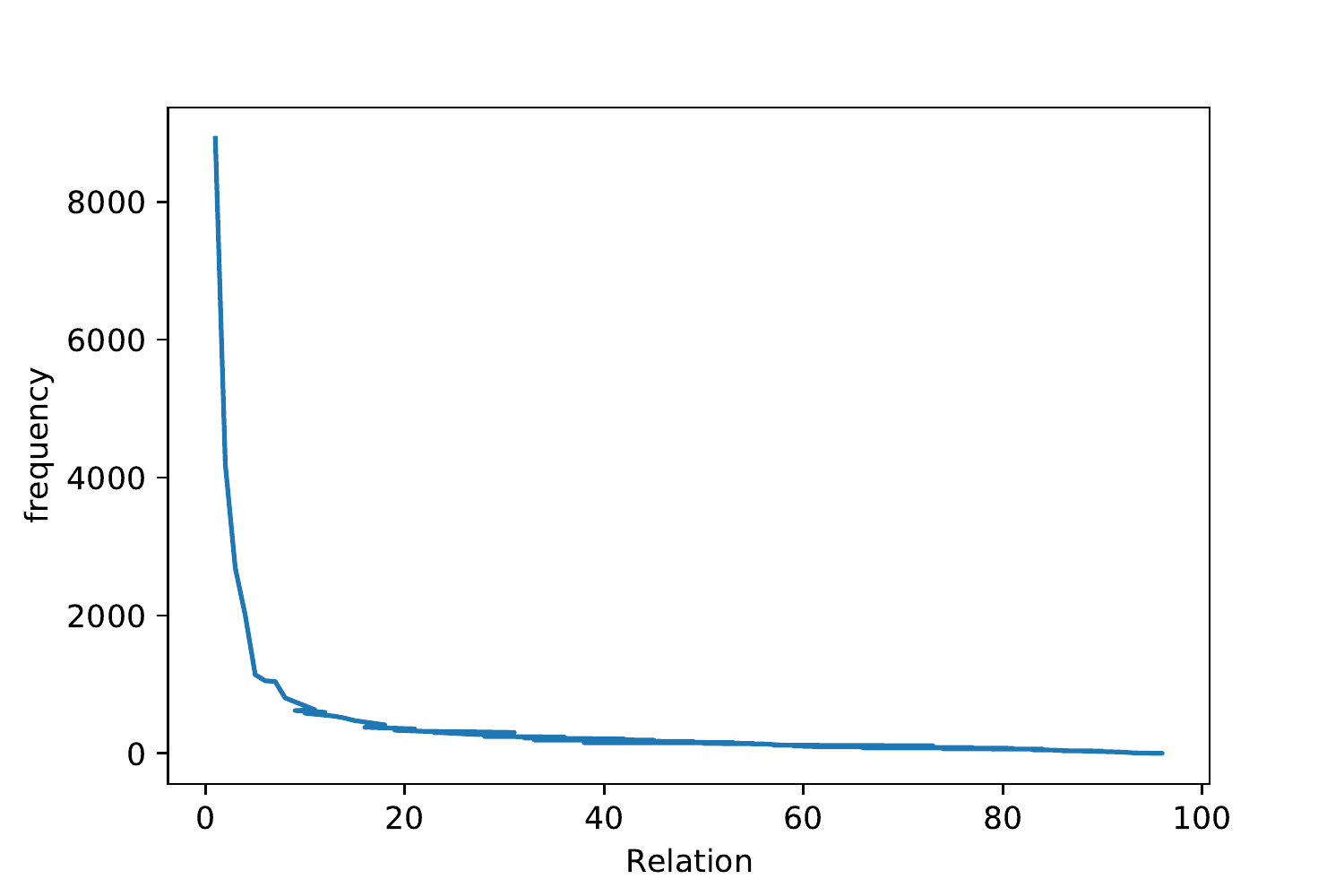}
    \caption{Relation Distribution of DocRED Train set. Relation index are sorted by frequency count from high to low.}
    \label{fig:docred-long-tail-distribution}
\end{figure}

Another unique challenge of DocRE that cannot be overlooked is long-tailed distribution. Long-tailed distribution is a common phenomenon in real-world data. In DocRE, we also observe the long-tailed distribution. Figure \ref{fig:docred-long-tail-distribution} presents the relation distribution of DocRED\citep{yao-etal-2019-docred}, a widely-used DocRE dataset: $7$ most frequent relations from $96$ relations takes up $55.12\%$ of total relation triples; while the frequencies of $60$ relations are only less than $200$. Vanilla training on long-tailed data will cause the model to achieve overwhelming performance on head relations but underfitting on tailed relations. Although the overall DocRE performance is largely dependent on performance on head relations since they are the majority, model failure on tailed relations is a big concern in real-world DocRE scenarios. 

Data augmentation is a commonly used strategy for addressing the long-tailed problem. Nonetheless, applying data augmentation efficiently on DocRE is non-trivial. Ordinary data augmentation operation on the document, including text random-dropping or replacing\citep{wei-zou-2019-eda} would require the DocRE model for extra encoding process of the entire document, which is computation in-efficient on DocRE since the document may contain numerous sentences. Besides, DocRE is a one-example-multi-instances task, so tailed relations and head relations presumably co-exist in one document. As a result, the head relations would also be augmented if we augment the tailed relations by aforementioned trivial augmentation methods on text, which is unexpected and may lead to over-fitting on head relations.

In this paper, we propose a novel data augmentation mechanism for DocRE, named ERA, for improving the document-level relation extraction by mitigating the long-tailed problem. The proposed ERA method applies augmentation on relation representations rather than texts, so it can augment tail relations without another encoding operation of the long document, which is computation-efficient and also effective for improving performance on tailed relations.

In addition, we propose a contrastive learning framework based on our ERA method, i.e., ERACL, for pre-training on the distantly-supervised data. The proposed ERACL framework can further enhance the model performance on tailed relations and achieve comparable overall DocRE performance compared to the state-of-art methods on DocRED.

\section{Background and Related Works}
\subsection{Problem Formulation}
Given a document $\mathcal{D}=\{w_1,w_2,...,w_l\}$ with $l$ words, a set of $n$ entities $\mathcal{E}= \{{e}_{i}\}^n_{i=1}$ are identified by human annotation or external tools. For each entity ${e}_i$, $m$ mentions of ${e}_i$ denoted as $\{m_{ij}\}^m_{j=1}$ are also annotated by providing the start position and end position in $\mathcal{D}$. In addition, the relation scheme $\mathcal{R}$ is also defined.

The objective of DocRE is to extract the relation triple set $\{(e_h,r,e_t)|e_h\in \mathcal{E},r\in \mathcal{R},e_t\in \mathcal{E}\} \subseteq \mathcal{E} \times \mathcal{R} \times \mathcal{E}$ from all possible relation triples, where each realtion triple $(e_h,r,e_t)$ extracted by the model can be interpreted as relation $r\in \mathcal{R}$ holds between head entity $e_h \in \mathcal{E}$ and tail entity $e_t \in \mathcal{E}$. For future simplicity, we denote tail relations as $\mathcal{R}^t \subset \mathcal{R}$ and head relations as $\mathcal{R}^h \subset \mathcal{R}$.
\subsection{Document-Level Relation Extraction}
\label{sec:docre-background}

To address the prior challenges in DocRE, one main branch of DocRE works use \textbf{Graph-based Methods}\citep{sahu_inter-sentence_2019,christopoulou_connecting_2019,wang_global--local_2020,zeng_double_2020,nan_reasoning_2020,li_graph_2020,xu_discriminative_2021}. The general idea of graph-based methods is to conduct multi-hop reasoning across entities, mentions and sentences in a document by graph neural networks. First a document is converted to a document graph by human designed heuristics, attention mechanism or dependency parser. Then the document graph is encoded by graph neural networks\cite{DBLP:conf/iclr/KipfW17,NEURIPS2018_182be0c5,wu2021graph} to conduct multi-hop reasoning across graph nodes and edges. Another branch of DocRE methods adopt \textbf{Transformer-based Methods}\citep{wang_fine-tune_2019,ye_coreferential_2020, DBLP:conf/aaai/XuWLZM21, zhou_atlop_2021, zhang_docunet_2021}. Transformer-based methods rely on the strong long-context representation capability of pre-trained transformers\citep{devlin-etal-2019-bert,liu_roberta_2019}. In addition, self-attention mechanism in transformer architecture can implicitly model the dependency between entities, mentions and contexts, which can be utilized for relation reasoning\citep{zhou_atlop_2021}. 



Different from previous works, in this paper we focus more on addressing the challenges of long-tailed distribution in DocRE. 
\subsection{Contrastive Learning}
\label{sec:contrastive-learning-background}
Contrastive learning is proved to be a promising self-supervised pre-training task for image recognition\citep{pmlr-v119-chen20j,he_momentum_2020}.
The principle of contrastive learning is to increase the representation similarity of anchor example $x$ and positive examples $x^+$ while decreasing the representation similarity of anchor example $x$ and negative examples $x^-$ by INFONCE loss\citep{Oord_INFONCE_2018}. 

Under the self-supervised setting, positive samples $x^+$ are constructed by data augmentation operation, including image cropping, resizing on anchor samples. The motivation of creating $x^+$ via data augmentation is that augmented samples are still similar or even the same in semantic space, then it can provide training signals for self-supervised pre-training. Therefore, models pre-trained by self-supervised contrastive learning can learn task-agnostic and robust representation for down-streaming tasks, which also can capture the semantic information of input samples.

The general contrastive learning framework has been applied in language tasks and achieved competitive performance.  \citet{fang_cert_2020} adapted the contrastive learning framework for self-supervised pre-training on transformers and achieved superior performance compared to BERT\cite{devlin-etal-2019-bert}. \citet{sclnlp_gunel_2021} proposed to use supervised contrastive learning for more robust fine-tuning on pre-trained transformers. In relation extraction, \citet{peng-etal-2020-learning} and \citet{qin-etal-2021-erica} adopted contrastive learning as one pre-training task to improve the relation understanding capability of BERT. It has been demonstrated that proper adaptation of contrastive learning framework can encourage the model to learn more robust task-agnostic or task-related representations, especially when the training data is limited. The problem of long-tail relation distribution is essentially lack of training samples of certain relation types. Considering this, we proposed a new contrastive learning framework based on a new data augmentation method in DocRE, called Easy Relation Augmentation(ERA), which can learn more robust relation representations for DocRE, especially for the tailed relations.

\section{Easy Relation Augmentation}
\subsection{Overview}
    We summarize the main components of ERA framework as follow: ERA takes a document $\mathcal{D}$ as input, then the Document Encoding and Relation Encoding modules will encode each entity pair $(e_h,e_t) \in \mathcal{E} \times \mathcal{E}$ from two aspects: contextualized entity representation and pooled context representation via self-attention mechanism of Pre-trained Transformers\citep{zhou_atlop_2021}. Afterwards, we proposed a novel Easy Relation Augmentation(ERA) mechanism to enhance the entity pair representation by applying a random mask on pooled context representation. The proposed ERA mechanism can augment the tail relations $r \in \mathcal{R}^t$ without another Relation Encoding and Document Encoding, which is computation-efficient and also effective. Finally, we train the relation prediction module on the augmented relation representations.
\subsection{Document Encoding}
    In light of the promising capability of Pre-trained Transformers\citep{devlin-etal-2019-bert,liu_roberta_2019} for modeling the long-range text dependency, We resort to pre-trained transformers for document encoding. We add a special entity marker "*"\citep{zhang-etal-2017-position} at the start and end position of each mention $m_{ij}$, and "*" can be replaced with other special tokens. Entity markers can spotlight the mention words and also provide entity positional information for Pre-trained Transformers, which proves to be effective in DocRE\citep{zhou_atlop_2021}. Feeding the document $\mathcal{D}$ to the pre-trained transformers, we can get the contextualized representation $\mathbf{H}$ of all words and vanilla multi-head self-attention $\mathbf{A}$ from the last block of Pre-trained Transformers($\mathbf{Ptr}$).
    \begin{equation}
        \mathbf{H},\mathbf{A} = \mathbf{Ptr}(\mathcal{D}=\{w_1,w_2,...,w_l\})
    \end{equation}
    Where $\mathbf{H} \in \mathbb{R}^{l \times d}$, $\mathbf{A} \in \mathbb{R}^{l \times l \times h}$. $d$ is the model dimension of the Pre-trained Transformers and $h$ is the number of self-attention heads of Pre-trained transformers.
\subsection{Relation Encoding}
    Given the contextualized representation $\mathbf{H}$ and self-attention $\mathbf{A}$ of the document $\mathcal{D}$, the goal of Relation Encoding module is to encode each entity pair $(e_h,e_t) \in \mathcal{E} \times \mathcal{E}$ by aggregating the contextualized entity representation and pooled context representation, which are crucial for relation understanding and reasoning across the long document.
    
    Contextualized entity representation can provide the contextualized entity naming and entity typing information for relation inference. For entity $e_h \in \mathcal{E}$, we obtain the contextualized mention representation by collecting the Pre-trained transformer last layer output of "*" marker at the start of mention $m_{ij}$, denoted as $\mathbf{m_{hj}}$. Subsequently, we can get the final contextualized entity representation $\mathbf{e_h}$ by logsumexp pooling \citep{jia-etal-2019-document}, which can achieve better results compared to max pooling and average pooling on DocRE\citep{zhou_atlop_2021}.
    \begin{equation}
        \mathbf{e_h} = log \sum _{j=1}^{m} exp(\mathbf{m_{hj}})
    \end{equation}
    
    As mentioned in Section \ref{sec:docre-background}, DocRE requires the model to capture the dependencies among entities, mentions, and context words, and also filter out the unnecessary context information from the long document. We named the aforementioned information as pooled context information. The self-attention matrix $\mathbf{A} \in \mathbb{R}^{l\times l \times h}$ obtained from Pre-trained transformers have already implicitly modeled the dependency among entities, mentions, and context words, which can be utilized for getting meaningful pooled context representation\citep{zhou_atlop_2021}. We follow \citet{zhou_atlop_2021} to obtain the pooled context information by utilizing the self-attention matrix $\mathbf{A}$.
    
    Given a entity pair $(e_h,e_t) \in \mathcal{E} \times \mathcal{E}$, one can get the pooled context representation $\mathbf{c_{h,t}}$ by Equation \ref{eq:pooled-context-representation} and \ref{eq:attention-dot-product}.
    \begin{equation}
        \label{eq:pooled-context-representation}
        \mathbf{c_{h,t} = H^T \cdot \frac{A_{h,t}}{1^T\cdot A_{h,t}}}
    \end{equation}
    
    \begin{equation}
        \label{eq:attention-dot-product}
        \mathbf{A_{h,t}} = \mathbf{A_h} * \mathbf{A_t}
    \end{equation}
    Where $\mathbf{A_h} \in \mathbb{R}^{l \times 1}$,$\mathbf{A_t} \in \mathbb{R}^{l \times 1}$ and $\mathbf{1} \in \mathbb{R}^{l \times l}$. $\mathbf{A_h}$ is the attention score of entity $e_h$ to all words in $\mathcal{D}$, which is obtained by averaging the attention score of all entity mentions $m_{hj}$, denoted as $\mathbf{A_{m_{hj}}}$. Similar to contextualized mention representation $\mathbf{m_{hj}}$, we obtain the mention attention score $\mathbf{A_{m_{hj}}}$ by indexing the vanila self-attention matrix $\mathbf{A}$ with position of starting "*" marker. In addition, note that the vanila self-attention matrix is first averaged over all attention heads before performing the indexing. $\mathbf{A_t}$ is also calculated following the same procedure.
    
    In the end, for the entity pair $(e_h,e_t)$, we can form a triple represention $\mathbf{\mathcal{T}_{h,t}} = \left( \mathbf{e_h},\mathbf{c_{h,t}},\mathbf{e_t} \right)$. $\mathbf{\mathcal{T}_{h,t}}$ contains all the information for relation prediction and form the basis for our Easy Relation Augmentation and Contrastive Leaning framework.
    
\subsection{Relation Representation Augmentation}
    \label{sec:era-relation-aug}
     To address the long-tailed problem residing in the DocRE, we propose a novel Easy Relation Augmentation(ERA) mechanism to increase the frequency of tailed relations and enhance the entity pair representation.
    
    
    Denote the set of triple representation of all entity pairs as $\mathcal{T}_{orig} = \{(\mathbf{e_h,c_{h,t},e_t}) |  e_h \in \mathcal{E}, e_t \in \mathcal{E}\}$. In addition, we can manually select the set of relations need to be augmented, i.e., $\mathcal{R}^{aug} \subseteq \mathcal{R}$. 
    
    Given a entity pair $(e_h,e_t)$ whose relation $r \in \mathcal{R}^{aug}$, we first retrieve the original triple representation $(\mathbf{e_h,c_{h,t},e_t})$ from $\mathcal{T}_{orig}$. Recall that the pooled context representation $\mathbf{c_{h,t}}$ encodes the unique context information for relation inference, and a slight perturbation on the context should not affect the relation prediction. Established on this intuition, we add a small perturbation on $\mathbf{c_{h,t}}$.
    
    We first apply a random mask on $\mathbf{A}_{h,t}$ described in Equation \ref{eq:pooled-context-representation} by multiplying $\mathbf{A}_{h,t}$ with a randomly generated mask vector $\mathbf{p} \in \mathbb{R}^{l\times 1}$. Each dimension of $\mathbf{p}$ is in $\{0,1\}$ and generated by a Bernoulli distribution with parameter $p$.
    \begin{equation}
        \mathbf{A'_{h,t}} = \mathbf{p} * \mathbf{A_{h,t}}
    \end{equation}
    Applying the random mask on attention score $\mathbf{A_{h,t}} \in \mathbb{R}^{l\times 1}$ can be interpreted as randomly filter out some context information since the attention score for them are set to $0$. In addition, the degree of perturbation can be controlled by setting proper $p$.
    Then we can get the perturbed pooled context representation $c'_{h,t}$ in Equation \ref{eq:pertubed-context-representation}.
    \begin{equation}
        \label{eq:pertubed-context-representation}
        \mathbf{c'_{h,t} = H^T \cdot \frac{A'_{h,t}}{1^T\cdot A_{h,t}}}
    \end{equation}
    
    For all the entity pairs $(e_h,e_t)$ whose relation $r$ in $\mathcal{R}^{aug}$, we apply the prior steps to get $\alpha$ distinct pertubed context representations $\{\mathbf{c'_{i,h,t}}\}^{|\alpha|}_{i=1}$ by using $\alpha$ random mask, where $\alpha$ is a hyper-parameter for controlling the number of ERA operations. Eventually, we can get the augmented triple representation set $\mathcal{T}_{aug}$, which can be formulated in Equation \ref{eq:augset-formulaiton}.
    \begin{equation}
        \label{eq:augset-formulaiton}
        \mathcal{T}_{aug} = \{\mathbf{(e_h,c'_{i,h,t},e_t)} | e_h \in \mathcal{E}, r \in \mathcal{R}^{aug}, e_t \in \mathcal{E}\}
    \end{equation}
    
    Combining the original triple representation set $\mathcal{T}_{orig}$ and $\mathcal{T}_{aug}$, we can get the total tripe representation set $\mathcal{T}$ for relation prediction and our Contrastive Learning framework.
    \begin{equation}
        \mathcal{T} = \mathcal{T}_{orig} \cup \mathcal{T}_{aug}
    \end{equation}
    
\subsection{Relation Prediction}
    Based on the triple representation of all entity pairs, the relation prediction module finally predict the relations hold between each pair. For a triple representation $\mathbf{(e_h,c_{ht},e_t) \in \mathcal{T}}$, we first apply two linear transformations with Tanh activation to fuse the pooled context representation $\mathbf{c}_{ht}$ with $\mathbf{e}_h$ and $\mathbf{e}_t$.
    \begin{equation}
        \label{eq:h_fusing}
        \mathbf{h} = tanh(\mathbf{W}_h \cdot \mathbf{e}_h + \mathbf{W}_{c1} \cdot \mathbf{c}_{h,t})
    \end{equation}
    \begin{equation}
        \label{eq:t_fusing}
        \mathbf{t} = tanh(\mathbf{W}_t \cdot \mathbf{e}_t + \mathbf{W}_{c2} \cdot \mathbf{c}_{h,t})
    \end{equation}
    Where $\mathbf{W}_{h},\mathbf{W}_{t},\mathbf{W}_{c1},\mathbf{W}_{c2} \in \mathbb{R}^{d\times d}$, which are trainable parameters of the model. Following \citet{zhou_atlop_2021}, then we used a grouped bi-linear layer to calculate a score for relation $r$, which splits the vector representation to $k$ groups and performs bilinear within group.
    \begin{equation}
        score_{r} = \sum_{i=1}^{k} \mathbf{h}^{iT}  \mathbf{W}_r^i \mathbf{t}^i
    \end{equation}
    Where $\mathbf{W}_r^i \in \mathbb{R}^{d/k \times d/k}$ is the bilinear parameter of group $i$. During training stage, we apply the adaptive thresholding loss \citep{zhou_atlop_2021} to dynamically learn a threshold $\theta_{h,t}$ for each entity pair by introducing a threshold class $TH$.
    \begin{equation}
        \label{eq:atlop-loss}
        \begin{aligned}
            \mathcal{L}_{h,t} = & -\sum _{r\in \mathcal{P}_{h,t}} log \left[ \frac{exp(score_r)}{\sum _{r' \in \mathcal{P}_{h,t} \cup \{TH\}} exp(score_{r'}) }\right]\\
                                & - log \left[ \frac{exp(score_{TH})}{\sum _{r' \in \mathcal{N}_{h,t} \cup \{TH\}} exp(score_{r'})}\right]
        \end{aligned}
    \end{equation}
    $\mathcal{P}_{h,t} \subset \mathcal{R}$ is the set of all valid relations that hold between entity pair $(e_h,e_t)$, and it is empty when no relation hold between the pair. In addition, $\mathcal{N}_{h,t} = \mathcal{R} - \mathcal{P}_{h,t}$. In the inference stage, the threshold $\theta$ for valid relation scores is set to $score_{TH}$.

\section{Contrastive Learning for relation pre-training}
\subsection{Overview}
   We propose a contrastive learning(CL) framework for unifying the augmented relation representations and improving the robustness of learned relation representations, especially for tailed relations. Specifically, we use the CL framework for pre-training on the distantly-supervised DocRE dataset\citep{yao-etal-2019-docred}, which is annotated by querying the knowledge graph but is noised. Considering that the model will be fine-tuned on the human-annotated dataset after the representation learning stage, the noise in the distantly supervised dataset is acceptable and correctable.
    
    Under DocRE setting, we claim that the semantically-similar samples should be the entity pairs that have the same relation $r$, including both of the original pairs and augmented pairs by ERA. However, only a few entity pairs have the same relation within one document, especially for the tailed relation $r$. 
    
\begin{figure}[t]
        \centering
        \includegraphics[width=8cm]{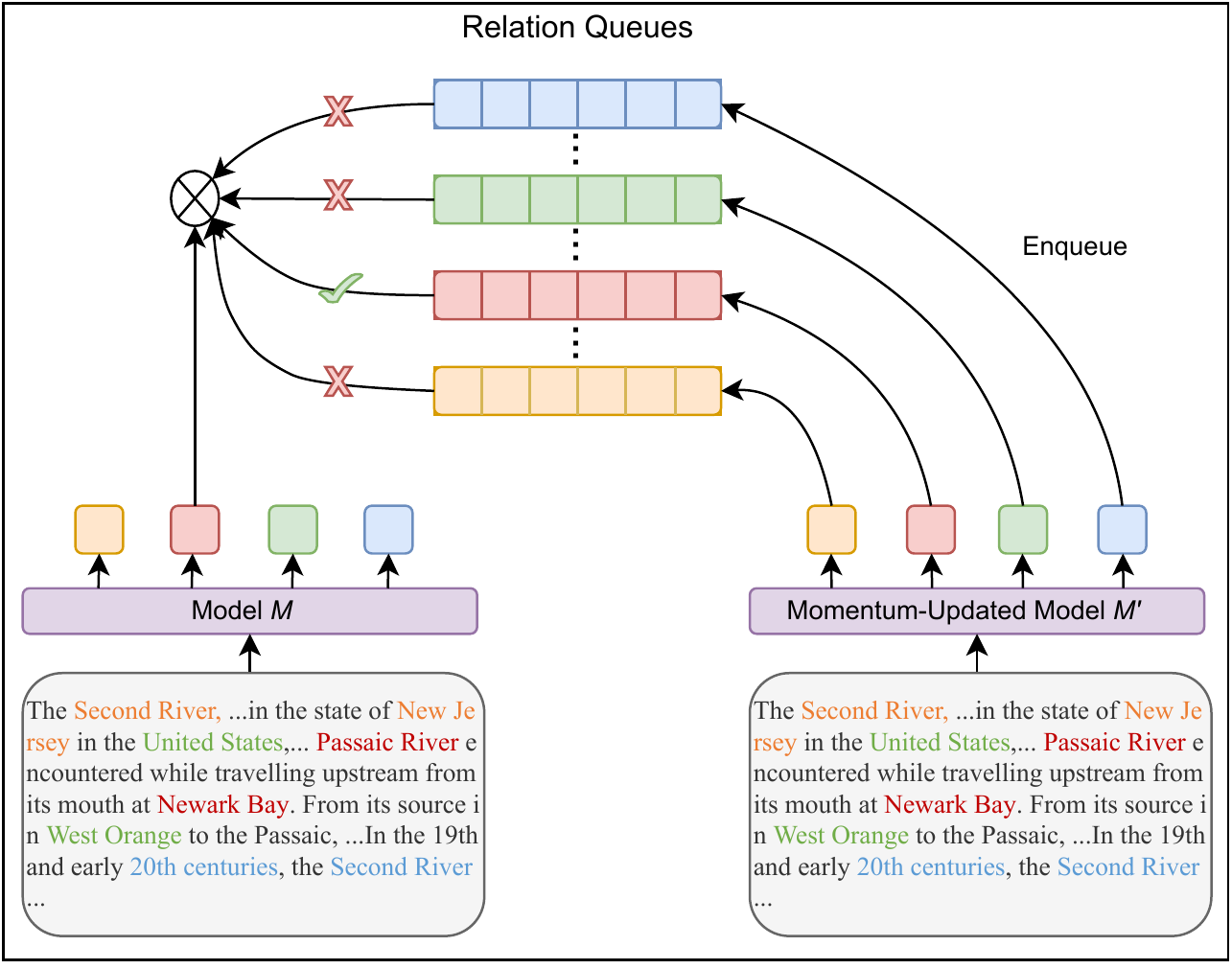}
        \caption{Overview of the proposed MoCo-DocRE framwork.}
        \label{fig:cl_overview}
    \end{figure}

    Increasing the mini-batch size can partially mitigate the problem, but it requires large GPU memory for training which may not be accessible. Thus, we adapted the MOCO framework \citep{he_momentum_2020} to the DocRE setting, named MoCo-DocRE. The moCo-DocRE framework can conduct the CL without using large batch-size by keeping a relation representation queue $Q_r$ holding $q$ relation representations from the previous mini-batch for each relation $r \in \mathcal{R}$. This allows us to reuse the encoded positive and negative relation representations in prior mini-batches. We summarize our contrastive learning framework in Figure \ref{fig:cl_overview}.
\subsection{Anchor relation encoding}
    For document $\mathcal{D}$ in pre-training dataset, we first conduct the aforementioned document encoding, relation encoding and Easy Relation Augmentation(ERA) and obtain the triple representation set $\mathcal{T}$ of all entity pairs. For a triple representation $\mathbf{(e_h,c_{h,t},e_t)} \in \mathcal{T}$, we use two linear transformations to fuse the triple representation, which are same as Equation \ref{eq:h_fusing} and \ref{eq:t_fusing}. Next, we use a MLP layer with ReLU activation for final relation representation:
    \begin{equation}
        \mathbf{x} = relu(\mathbf{W}_2(\mathbf{W}_1 [\mathbf{h:t}] + \mathbf{b}_1) + \mathbf{b}_2)
    \end{equation}
    Where $[:]$ denotes the vector concatenation operation, $\mathbf{W}_1 \in \mathbb{R}^{2d\times d}$ and $\mathbf{W}_2 \in \mathbb{R}^{d\times d_r}$ are trainable model parameters in pre-training stage, and $d_r$ is the dimension of final relation representation $\mathbf{x}_{h,t}$. After contrastive pre-training, the MLP layer will not be used for relation prediction in fine-tuning.
\subsection{MoCo-DocRE}
    To keep the consistency of relation representation in $Q_r$, we also use a momentum updated model to encode the positive and negative samples in contrastive learning\citep{he_momentum_2020}. The original model $\mathcal{M}$ is updated via back-propagation, and the momentum-updated model $\mathcal{M}'$ is updated by Equation \ref{eq:momentum-updated}. 
    \begin{equation}
        \label{eq:momentum-updated}
        \mathcal{M}' = m \cdot \mathcal{M}' + (1-m) \cdot \mathcal{M}
    \end{equation}
    Where $m$ is the momentum hyper-parameter, which can control the evolving speed of $M'$. Next we feed the document $\mathcal{D}$ to $\mathcal{M}'$ for $\mathbf{x}'$ by following the same procedure as getting anchor relation representation $\mathbf{x}$. Then we push $\{\mathbf{x'} | (\mathbf{e}_h,\mathbf{c}_{h,t},\mathbf{e}_t) \in \mathcal{T}'\}$ to $|\mathcal{R}|$ relation representation queues according to their relation labels. If relation $r$ holds between $(e_h,e_t)$, then $x'$ will be pushed to $Q_r$. Eventually, we can get the set of positive and negative relation representations of $\mathbf{x}$ from queues, i.e, $\mathcal{P} = \cup _{r \in \mathcal{P}_{h,t}} Q_r$ and $\mathcal{N}=\cup _{r \in \mathcal{N}_{h,t}} Q_r$.
    
    For anchor relation representation $\mathbf{x}$, now we can formalize the INFONCE loss \citep{Oord_INFONCE_2018} under our MoCo-DocRE in Equation \ref{eq:info-nce}.
    \begin{equation}
        \label{eq:info-nce}
        \mathcal{L} = - \sum_{x^+ \in \mathcal{P}} log \left[ \frac{e^{\mathbf{x}^T  \mathbf{x^+ / \tau}}}{e^{\mathbf{x}^T  \mathbf{x^+/\tau}} + \sum_{x^- \in \mathcal{N}}{e^{\mathbf{x}^T \mathbf{x}^-/\tau}}}   \right]
    \end{equation}
    Where $\tau$ is the temperature hyperparameter. In addition $\mathbf{x},\mathbf{x}^+,\mathbf{x}^-$ in Equation \ref{eq:info-nce} are \textit{l2}-normalized.

\begin{table*}[ht]
    \centering
    \begin{tabular}{lcccc}
        \toprule
        \multirow{2}*{\textbf{Model}} & \multicolumn{2}{c}{\textbf{Dev}} & \multicolumn{2}{c}{\textbf{Test}}  \\ 
         & Ign $F_1$ & $F_1$ & Ign $F_1$ & $F_1$ \\
        \midrule
        \textit{Graph-based Methods}  & & & &\\     
        GLRE\citep{wang_global--local_2020} & -- & -- & 55.40 & 57.40 \\
        LSR\citep{nan_reasoning_2020} &52.43&  59.00& 56.96& 59.05 \\
        GAIN\citep{zeng_double_2020} &59.14& 61.22 & 59.00& 61.24\\
        DRN\citep{xu_discriminative_2021} &59.33& 61.39& 59.15& 61.37 \\
        \midrule
        \textit{Transformer-based Methods} & & & &\\ 
        SSAN\citep{DBLP:conf/aaai/XuWLZM21} &57.04& 59.19& 56.06& 58.41 \\
        ATLOP\citep{zhou_atlop_2021} & 59.22  & 61.09  &59.31 &61.30 \\
        DocuNet\citep{zhang_docunet_2021} &59.86 &61.83& 59.93& 61.86 \\
        AFLKD\citep{tan-etal-2022-document} &60.08 &62.03 &60.04 &62.08 \\

        \midrule
        \textit{Our Methods} & & & &\\
        ERA & 59.30 $\pm$ 0.09 & 61.30 $\pm$ 0.08 &58.71 &60.97 \\
        ERACL &59.72 $\pm$ 0.19 & 61.80 $\pm$ 0.20&59.08 &61.36  \\
        \toprule
    \end{tabular}
    \caption{Overall DocRE performance evaluated on DocRED benchmark. We report the mean and standard deviation of 3 runs with different random seeds on the development set. The official test results are reported by the best checkpoint on the development set. Results of all other models are reported in their original paper and use BERT-base-cased as backbone encoder.}
    \label{tb:main-results}
\end{table*}

\section{Experiment}
\subsection{Experiment Setup and Dataset}
   \begin{table}[h]
        \centering
        \begin{tabular}{lcc}
            \toprule

            Statistics & DocRED & HacRED \\
            \midrule
            \# Train & 3053 &6231 \\
            \# Dev &1000 &1500 \\
            \# Test &1000 &1500 \\
            \# Relations &96 &26 \\
            \toprule
        \end{tabular}
        \caption{Dataset Statistics of DocRED and HacRED}
          \label{tb:data-stat}
    \end{table}
    
    \textbf{Dataset:}
   We evaluate the proposed ERA and contrastive learning framework on two popular DocRE datasets, DocRED\citep{yao-etal-2019-docred} and HacRED\citep{cheng-etal-2021-hacred}. DocRED contains 5053 English documents extracted from Wikipedia and 96 relations, which are human-annotated. Besides, DocRED also provide a distantly-supervised dataset with 101873 documents, and the relation of entitity pairs are annotated by querying Wikidata. HacRED is a human annotated Chinese dataset with 26 relations. Statistics of Datasets are listied on Table \ref{tb:data-stat}.  \\
    \textbf{Implementation Details:} We use the pre-trained BERT-base-cased\citep{devlin-etal-2019-bert} as our backbone for DocRED dataset. All the hyperparameters are tuned on the development set. Specifically, we set the random mask probability $p$ to 0.1 and the number of augmentation $\alpha$ to $2$. In addition, the number of grouped bilinear $k$ is set to $64$. The temperature parameter $\tau$ is set to $0.5$ and the size of $Q_r$, i.e., $q$ is set to $500$, and the momentum $m$ is set to 0.99. The learning rate is set to $1e-5$ for pre-training on our CL framework. In the fine-tune on human-annotated data, we set the learning rate to $5e-5$ for parameters of BERT and $1e-4$ for other parameters. We use AdamW\citep{losch_adamw_2019} for optimization of all parameters and a linear-decayed scheduler with a warmup ratio $0.06$. Gradients whose norm is larger than $1$ are clipped. For HacRED dataset, we use XLM-Roberta-base\citep{conneau-etal-2020-unsupervised} as backbone. Under HacRED scenario, we set the random mask probability $p$ to 0.05 and the number of augmentation $\alpha$ to $3$. All the other parameters are same as the DocRED scenario. 
    
\subsection{Evaluation Metric}
    DocRED benchmark provide two evaluation metrics, i.e. $F_1$ Ign $F_1$. $F_1$ is the minor $F_1$ value for all predicted relations in test/development dataset, which can reflect the overall performance of DocRE. Compared to $F_1$, Ign $F_1$ excludes the entity pairs which appear both on training and test/dev data.
    To demonstrate how ERA and contrastive learning can improve the performance of tailed relations, we propose to use the following evaluation metrics: \\
    $\mathbf{Macro}$: it computes the $F_1$ value by first calculating $F_1$ for each relation separately and then getting the average of all relation classes. Compared to minor $F_1$, macro $F_1$ treat all relation classes equally, $F_1$ of tailed relations will have equal impact compared to head relations.\\
    $\mathbf{Macro@500}$,$\mathbf{Macro@200}$,$\mathbf{Macro@100}$: Those metrics target at tailed relations whose frequency count in train dataset is less than 500,200,100 respectively. Values are computed by averaging the F1 value of the targeted relations.

\begin{table*}[h]
\centering
    \begin{tabular}{lcccc}
    \toprule
    Methods & $\mathbf{Macro}$ & $\mathbf{Macro@500}$ & $\mathbf{Macro@200}$ & $\mathbf{Macro@100}$  \\
    \midrule
    DocuNet &40.69 & 36.54 & 28.95 & \bf{22.37}          \\
    ATLOP   & 39.54 & 35.20 & 26.82 & 18.76         \\
    \hline
    --Deletion &39.22 &34.88 &26.62 &18.78 \\
    --Mask &38.31 &33.80 &25.30 &17.35 \\
    --AFL &40.04 &35.78 &27.78 &20.52 \\
    \hline
    ERA     & 40.55 & 36.21 & 28.51 & 20.50 \\
    ERACL   & \bf{41.34} & \bf{37.13} & \bf{29.43} & 22.31     \\
    \toprule
\end{tabular}
    \caption{Evaluation on tailed relations. All the results are averaged on 3 runs with different random seeds on development set. Relation labels of test set are not accessible, so the results on test set cannot be reported.}
    \label{tb:tail-results}
\end{table*}

\subsection{Main Results}
We compare the proposed ERA and ERACL methods to graph-based and transformer-based models on the DocRED benchmark by using $F_1$ and Ign $F_1$ metrics on the dev/test dataset. Results are reported in Table \ref{tb:main-results}. The proposed ERACL method, which first conducts contrastive learning under our MoCo-DocRE framework on the distantly supervised dataset and then conducts ERA fine-tuning on the training set, can achieve competing $F_1$ and Ign $F_1$ value, compared to state-of-art graph-based methods and transformer-based methods. Besides, compared to ATLOP\citep{zhou_atlop_2021} which is the baseline of ERACL, ERACL can improve the minor $F_1$ on the development set by $0.71$ and ERA can improve the minor $F_1$ by $0.29$, which demonstrate the effectiveness of the proposed ERA method and contrastive learning pretraining.

\subsection{Results on tailed relation}
    To demonstrate the effectiveness of our ERA and ERACL on improving model performance on tailed relations, we evaluated ERA and ERACL using Macro, Macro@500, Macro@200, and Macro@100 metrics. Besides, we compare ERA and ERACL with three baseline methods which are used for addressing long-tailed distribution on DocRED. 
    \textbf{Text Random Deletion/Mask} are commonly used data augmentation techniques for NLP tasks. We apply the Text Random Deletion and masking as data augmentation for documents which contain  tailed relations. 
    \textbf{Adaptive Focal Loss} proposed by \citet{tan-etal-2022-document} is a adaptation of Focal Loss\citep{Lin_2017_ICCV} on DocRE scenario. AFL proved to be effective on tail relations. We implement those three methods based on ATLOP. Results are listed on Table \ref{tb:tail-results}. The proposed ERA and ERACL method can outperform those three baselines on tailed relations, which demonstrates the effectiveness and necessities of our ERA and ERACL for addressing the long-tailed distribution on DocRE scenario.
    
    Moreover, the proposed ERA method can improve the Macro over ATLOP by $1.01$, $1.01$ on Macro@500, $1.69$ on Macro@200, and $1.74$ on Macro@100. We observe that the improvements are more significant on relations that appear less frequently. In addition, the proposed ERACL method can further gain improvements over ERA: 0.79 on Macro, 0.92 on Macro@500, 0.92 on Macro@200, 1.81 on Macro@100, which also show similar trends as ERA over ATLOP.
    
    \begin{figure}[h]
        \centering
        \includegraphics[width=7cm]{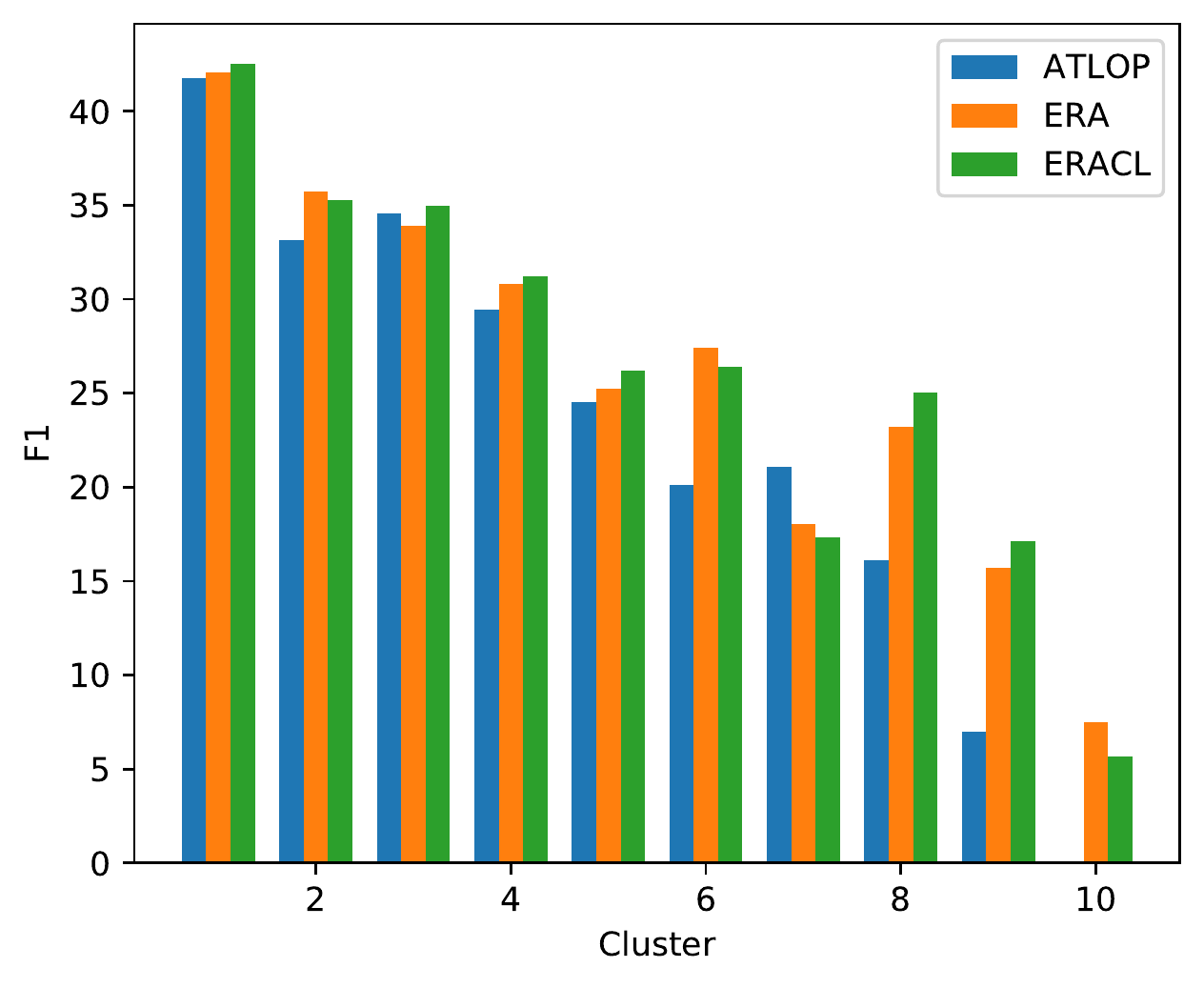}
        \caption{Comparison of F1 across 10 relation clusters. All results are averaged by 3 runs with different random seeds.}
        \label{fig:relationF1-demo}
    \end{figure}
    
      \begin{table*}[h]
        \centering
        \begin{tabular}{lccccc}
        \toprule
        Methods & F1 & $\mathbf{Macro}$ & $\mathbf{Macro@500}$ & $\mathbf{Macro@200}$ & $\mathbf{Macro@100}$ \\
        \midrule
        ERACL   & 61.80 & 41.34 & 37.13     & 29.43     & 22.31     \\
        \hline
        -- ERA   & 61.36 & 40.49 & 36.22     & 28.61     & 21.52     \\
        -- CL    & 61.30 & 40.55 & 36.21     & 28.51     & 20.50     \\
        -- both  & 60.97 & 39.54 & 35.20     & 26.82     & 18.76    \\
        \toprule
        \end{tabular}
        \caption{Ablation Study on development set.}
        \label{tb:ablation-study}
    \end{table*}
    
    To better illustrate the performance gain on the tailed relations, we sort 96 relations according to their frequency count in the DocRED train set from high to low, then slice 96 relations to 10 relation clusters equally for more clear visualization. For each cluster, we calculate the cluster F1 by averaging the F1 of relation within the cluster. The results are demonstrated in Figure \ref{fig:relationF1-demo}. We observe that the proposed ERA method gain improvements compared to ATLOP on relation clusters 4-10, which correspond to the tailed relation in DocRED, and also achieve competing performance on clusters 1-3, which correspond to the head relations. Those findings show that our ERA methods are effective for improving the DocRE performance on tailed relations while keeping the performance on head relations. In addition, similar performance gain is also achieved by the proposed ERACL method, and ERACL can further improve the tailed relations compared to ERA and achieve competing performance on head relations. 
    
    In addition, we conduct another set of experiments by manually reducing the percentage of training data in order to explore the performance of the proposed ERA methods and ERACL methods under a limited-data scenario. The results are listed in Table \ref{tb:with-few-data}. Compared to the setting that uses all of the train data, we observe that the performance gain of the proposed ERA and ERACL under 10\% and 5\% settings are more significant, which also indicate that the proposed ERA and ERACL can improve the DocRE performance by mitigating the long-tailed problem and are especially effective when training data is limited.
    \begin{table}[!h]
        \centering
        \begin{tabular}{lccc}
            \toprule
            Methods & F1 & $\mathbf{Macro}$ & $\mathbf{Macro@200}$\\
            \midrule
            \textit{10\%} &  &  &     \\
            ATLOP &51.92   &23.68  &7.97  \\
            ERA    &52.46   &24.80  &11.20   \\
            ERACL    &\bf{53.73}   &\bf{27.88}  &\bf{14.89}   \\
            \hline
            \textit{5\%} &  &  &     \\
            ATLOP    &45.17   &16.17   &5.59    \\
            ERA     &47.18  &17.27    &6.22    \\
            ERACL     &\bf{49.40}   &\bf{23.47}   &\bf{11.84}    \\
            \toprule
        \end{tabular}
        \caption{Results on the development set under the limited-data setting. 10\% refers to only using 10\% of training data, and 5\% refers to only using 5\% of training data. All results are reported by averaging 3 runs with different random seeds.}
        \label{tb:with-few-data}
    \end{table}
    
    Besides, we also conduct experiments on HacRED to investigate whether our ERA can generalize well on other long-tailed DocRE datasets. Results are shown on Table \ref{tb:hacred-results}. We observe that ERA can still outperform the ATLOP on tailed relations.
    
     \begin{table}[!h]
        \centering
        \begin{tabular}{lccc}
            \toprule
            Methods & F1 & $\mathbf{Macro}$ & $\mathbf{Macro@500}$\\
            \midrule
            ATLOP &77.84   &70.99  &55.11  \\
            ERA    &78.27   &71.73  &57.13   \\
            \hline
            \toprule
        \end{tabular}
        \caption{Results on HacRED. Since HacRED do not have distant-labeled data, we can only evaluate ERA on HacRED. ATLOP results are implemented by us. All experiments use XLM-Roberta-base as the backbone encoder.}
        \label{tb:hacred-results}
    \end{table}

\subsection{Ablation Study}
    To evaluate the contribution of the ERA and contrastive learning(CL) framework separately, we conduct an ablation study on the development set by reducing one component at a time. The results are shown in Table \ref{tb:ablation-study}. All of the results are tuned on the development set for best performance. 
    
    Note that reducing ERA refers to turning off the relation representation augmentation operation described in Section \ref{sec:era-relation-aug} and only keeping the original relation representations. In addition, reducing CL means without conducting contrastive learning on distantly supervised data. We observe that the ERA component and contrastive learning(CL) framework are almost equally important, which lead to $0.44$ and $0.50$ performance drop on F1 metric, $0.85$ and $0.79$ performance drop on Macro F1.
    
\section{Conclusion}
   We propose a novel Easy Relation Augmentation(ERA) method for the Document-level Relation Extraction task, which improves the DocRE performance by addressing the long-tailed problem residing in DocRE by augmentation on relation representations. In addition, we propose a novel contrastive learning framework based on ERA, i.e., MoCo-DocRE, for unifying the augmented relation representations and improving the robustness of learned relation representations, especially for tailed relations. Experiments on the DocRED dataset demonstrate that the proposed ERA and ERACL can achieve competing performance compared to state-of-arts models, and we demonstrate that the performance gain of ERA and ERACL are mainly from the tailed relations.
    
    Nonetheless, addressing the long-tailed problem is still challenging for DocRE. One limitation of our method is it still relies on large amount of annotated data to achieve overwhelming performance. We hope it can be mitigated in future research.
\bibliography{anthology,custom}
\bibliographystyle{acl_natbib}




\end{document}